\documentclass{article}
\usepackage{ijcai25}

\usepackage{times}
\usepackage{soul}
\usepackage{url}
\usepackage[hidelinks]{hyperref}
\usepackage[utf8]{inputenc}
\usepackage[small]{caption}
\usepackage{graphicx}
\usepackage{amsmath}
\usepackage{amsthm}
\usepackage{booktabs}
\usepackage{algorithm}
\usepackage{algorithmic}
\usepackage[switch]{lineno}
\usepackage{amssymb}

\usepackage{lipsum}

\usepackage{multirow}
\usepackage{diagbox}

\newtheorem{definition}{Definition}

\title{Learning Counterfactually Fair Models via Improved Generation with Neural Causal Models}

\author{
Krishn V. Kher$^1$
\and
Saksham Mittal$^1$\and
Aditya Varun V$^1$\and
Shantanu Das$^{2}$\and 
SakethaNath Jagarlapudi$^1$\\
}

\begin{document}

\maketitle

\begin{abstract}
One of the main concerns while deploying machine learning models in real-world applications is fairness. Counterfactual fairness has emerged as an intuitive and natural definition of fairness. However, existing methodologies for enforcing counterfactual fairness seem to have two limitations: (i) generating counterfactual samples faithful to the underlying causal graph, and (ii) as we argue in this paper, existing regularizers are mere proxies and do not directly enforce the exact definition of counterfactual fairness. In this work, our aim is to mitigate both issues. Firstly, we propose employing Neural Causal Models (NCMs) for generating the counterfactual samples. For implementing the abduction step in NCMs, the posteriors of the exogenous variables need to be estimated given a counterfactual query, as they are not readily available. As a consequence, $\mathcal{L}_3$ consistency with respect to the underlying causal graph cannot be guaranteed in practice due to the estimation errors involved. To mitigate this issue, we propose a novel kernel least squares loss term that enforces the $\mathcal{L}_3$ constraints explicitly. Thus, we obtain an improved counterfactual generation suitable for the counterfactual fairness task. Secondly, we propose a new MMD-based regularizer term that explicitly enforces the counterfactual fairness conditions into the base model while training. We show an improved trade-off between counterfactual fairness and generalization over existing baselines on synthetic and benchmark datasets.%
\end{abstract}

\section{Introduction}\label{sec:intro}
\begin{figure} \centering \includegraphics[width=0.5\linewidth]{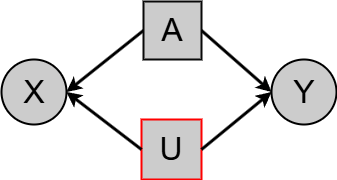} \caption{Causal graph for generating predictands using sensitive data. The variables outlined in black are observed attributes while the one in red is unobserved.} \label{fig:causalgraph} \end{figure}

As machine learning systems increasingly address real-world predictive tasks across diverse domains such as healthcare \cite{causalMLhealth,causalMLhealthSurvey}, econometrics \cite{causalMLEcon}, and climate change \cite{causalMLClimate,10.1145/3485128} the necessity of designing these systems to be free from unwarranted biases has become paramount. Ensuring fairness for all intended users requires more than merely evaluating predictive (training and generalization) error, as datasets often contain unequal proportions of data attributes or categories. Without explicit constraints promoting fairness, learning algorithms or models may inadvertently exploit these imbalances, minimizing predictive error on overrepresented samples while neglecting underrepresented ones, thereby maintaining low overall predictive error at the expense of fairness.
To rigorously assess fairness, numerous approaches and notions have been introduced, encompassing individual and group fairness settings, among others \cite{10.1145/3616865}. In this paper, we focus on a specific fairness criterion known as Counterfactual Fairness, initially proposed by \cite{10.5555/3294996.3295162} within the framework of individual fairness. Counterfactual Fairness is grounded in a causal model of the predictive task \cite{10.1145/3501714.3501755}, which delineates protected or sensitive attributes ($A$), other attributes ($X$), and the predictand ($Y$). Additionally, unobserved confounders ($U$) may be present and are typically treated as latent variables due to the lack of explicit supervision. An illustrative example of such a causal model, which we adopt throughout this paper, is presented in Figure \ref{fig:causalgraph}.

The fundamental principle of Counterfactual Fairness is that predictions from a model $h_{\phi}(\cdot,\cdot)$ should remain equitable with respect to any data contained in the sensitive attributes $A$ and any information in $X$ that may implicitly allude to $A$.
\begin{definition}\label{def:ctfairness}
    \textit{Counterfactual Fairness} \cite{10.5555/3294996.3295162}: A predictive model $h_\phi(\cdot, \cdot)$ is counterfactually fair if:
    \begin{align}
        \forall (x, y) \in \Omega_X \times \Omega_Y, P(h_\phi(x, a) = y|X=x, A=a) &= \nonumber \\ P(h_\phi(x, a') = y|X=x, A=a), \forall (a, a') \in \Omega_A \times \Omega_A. 
    \end{align}
\end{definition}
Despite significant advancements, existing works \cite{10.1007/s10994-022-06206-8,10.5555/3294996.3295162,10.1145/3580305.3599408,zuo2022counterfactual,anthis2023causal} suffer from two primary limitations:
\begin{enumerate}
    \item Generation of counterfactual samples that only partially adhere to the underlying causal graph;
    \item As we argue in this paper, most existing regularizers being mere proxies that do not directly enforce the exact definition of counterfactual fairness.
\end{enumerate}
While the first limitation is relatively less acute, since existing approaches typically involve explicit training for generating counterfactual data augmentation, the second limitation poses a more critical challenge. To address the former issue, we propose leveraging Neural Causal Models (NCMs).
NCMs \cite{10.5555/3540261.3541089,xia2023neural,Xia_Bareinboim_2024}, are a class of Structural Causal Models (SCMs) \cite{judeaclassic}, where the node-wise mechanisms are modeled using neural networks. Being a class of SCMs, they are proven to be $\mathcal{L}_3$ consistent with respect to the underlying causal graph (they satisfy level-$3$ constraints in the ladder of causal hierarchy), and are sufficiently expressive owing to the universal approximation capabilities of neural networks, (cf.  Theorem~\textbf{1, 2} in \cite{xia2023neural}). Consequently, they are well-suited for counterfactual inferences and generation.

However, for a given counterfactual query, the posterior distribution of the exogenous variables, essential for implementing the abduction step, is not readily available in an NCM. Accurately modeling and estimating this posterior is therefore necessary. Existing methodologies either restrict themselves to discrete variables \cite{xia2023neural} or assume invertibility of the node-wise mechanisms, either explicitly \cite{10.5555/3618408.3619478} or implicitly via encoder-decoder models \cite{chao2024modeling,NEURIPS2020_0987b8b3,10.24963/ijcai.2024/907}. Even when the true posterior satisfies these restrictions (i.e., with zero modeling error), finite-sample estimation errors can impede $\mathcal{L}_3$ consistency in practice. In this paper, we circumvent restrictive modeling assumptions by proposing to learn the posterior using a neural conditional generator, which is known to be universal \cite{pmlr-v125-kidger20a} (see Lemma 2.1 in \cite{liu2021wasserstein}). We employ a kernel least squares loss to train the parameters of this generator, aligning the model's posterior with that implicit in the data. By selecting appropriate kernels, our approach can handle discrete, continuous, or mixed data types. To mitigate inconsistency issues arising from finite samples, we introduce a novel data-driven loss term that explicitly enforces the $\mathcal{L}_3$ constraints. Importantly, this loss term can also enhance existing methodologies, potentially serving as an independent contribution.

Furthermore, most existing causal models based on deep generative networks are $\mathcal{L}_3$ identifiable if and only if the true SCM is so \cite{10.24963/ijcai.2024/907}. To practically address the vast majority of non-identifiable cases, we incorporate a regularizer that biases the model towards plausible counterfactual distributions. Specifically, our regularizer enforces a near-world assumption on the counterfactual distribution, ensuring that the counterfactual remains as close as possible to the factual evidence. Although similar regularizers have been utilized in the causal literature \cite{JMLR:v25:21-1440,torous2024optimaltransportapproachestimating}, their application within the context of NCMs appears novel. Empirical observations indicate that this regularizer produces counterfactual samples that align more closely with intuitive expectations.

Finally, regarding fairness, we critically observe that most prior works \cite{10.1007/s10994-022-06206-8,10.5555/3294996.3295162,zuo2022counterfactual} assess counterfactual fairness using weak metrics such as (R)MSE/MAE, which may not fully adhere to Definition \ref{def:ctfairness}, as discussed in Section \ref{subsec:ff}. We identify this limitation and propose a metric based on kernel means, which accurately determines whether the distributions in question are identical. Thus, our novel contributions are three-fold. We conclude by comparing our proposed method using our novel metric against a relevant baseline close to our setting \cite{10.1007/s10994-022-06206-8} and observe improved results.


\section{Background}\label{sec:background}
We begin by laying out certain pre-requisites in causality theory (we borrow the notation of \cite{10.5555/3540261.3541089} in this regard) and counterfactual fairness, followed by analyzing prior work that has attempted to tackle counterfactual fairness under varied settings. In general we denote random variables by uppercase letters ($W$) and their corresponding values by lowercase letters ($w$). We denote by $\mathcal{D}_{W}$ the domain of $W$ and by $P(W=w)$ the probability of $W$ taking the value $w$ under the probability distribution $P(W)$. We denote by $\Omega(W)$ the domain of values for a random variable $W$. Bold font on a semantic letter indicates a set. 

\subsection{Preliminaries}\label{subsec:prelims}
\textbf{Structural Causal Model} An SCM $\mathcal{M}$ is defined as a tuple $\langle \mathbf{U},\mathbf{V},\mathcal{F},P(\mathbf{U})\rangle$, where $U$ is a set of exogenous variables with distribution $P(\mathbf{U})$, $\mathbf{V}$ the endogenous variables, and $\mathcal{F}\equiv\{f_1,\ldots,f_n\}$ is a set of functions/mechanisms, where each $f_i:U_i\times A_i\rightarrow V_i, U_i\subset \mathbf{U},A_i\subset \mathbf{V}\setminus\{V_i\}$. In simple words, $\mathbf{V}$ are the observed variables, $\mathbf{U}$ are the unobserved variables, and each mechanism takes as input some subset of exogenous and endogenous variables and outputs the corresponding observed variable. So $U_i,A_i$ are the causes for the effect $V_i$. Given a \emph{recursive} SCM, a directed graph is readily induced: the nodes correspond to the endogenous variables $V$, while the directed edges correspond to the causal mechanism from each variable in $A_i$ to $V_i$. It is assumed that this directed graph is acyclic and as such, $A_i$ are the parents of $V_i$ in this DAG. Every causal diagram is associated with a set of $\mathcal{L}_1$ constraints, which is a set of conditional independences, $\mathcal{L}_2$ constraints, popularly known as the do-calculus rules for interventions and finally, $\mathcal{L}_3$ constraints, rules that the counterfactual distributions satisfy. Every counterfactual distribution induced by the SCM satisfies all three levels of constraints. Using these 3-layered rules, various statistical, interventional, and counterfactual inferences can be performed in a systematic way. Notably, distributions in the lower levels of the causal hierarchy may not satisfy constraints of higher levels. Critically, marginalizing over the posterior of a counterfactual distribution lends the interventional distribution under the corresponding intervention. This law is used in enforcing $\mathcal{L}_3$ consistency, as we elaborate later.
Since in most applications the cause-effect relations are known rather than the SCM itself, one interesting modeling question is, given a causal graph, can we come up with a convenient model that is consistent with all the three levels of constraints induced by the graph?
NCMs are one such example of a convenient family of causal models. 
\begin{definition}\label{defn:ncm}
\textit{NCM}, Defn.$2$ in \cite{Xia_Bareinboim_2024} 
Given a causal diagram $\mathcal{G}$, a $\mathcal{G}$-constrained NCM $\widehat{\mathcal{M}_{\theta}}$ over $\mathbf{V}$ with parameters $\mathbf{\theta} = \{\theta_{V_i} : V_i \in \mathbf{V}\}$ is an SCM $\langle \widehat{\mathbf{U}},\widehat{\mathbf{V}},\widehat{\mathcal{F}},P(\mathbf{\widehat{U}})\rangle$ such that (1) $\mathbf{\widehat{U}} =
\{\widehat{U}_\textbf{C} : \textbf{C} \in \mathbb{C}(\mathcal{G})\}$, where
$\mathbb{C}(\mathcal{G})$ is the set of all maximal cliques over bi-directional edges of
$\mathcal{G}$; (2) $\mathcal{\widehat{F}} = \{\widehat{f_{V_i}} : V_i \in \mathbf{V}\}$, where each $\widehat{f_{V_i}}$ is a feedforward neural net
parameterized by $\theta_{V_i} \in \mathbf{\theta}$ mapping $\mathbf{U}_{V_i} \cup \mathbf{A}_{V_i}$ to $V_i$ for $\mathbf{U}_{V_i} = \{
\widehat{U}_{\mathbf{C}} \in \widehat{\mathbf{U}} : V_i \in \mathbf{C}\}$ and $\mathbf{A}_{V_i}
= A_{\mathcal{G}}(V_i)$; (3) $\texttt{Unif}(0,1) \mapsto P(\widehat{U}), \forall~\widehat{U} \in \widehat{\mathbf{U}}$.
\end{definition}
Note that point $3$ in Definition \ref{defn:ncm} above critically lies on the fact that there always exists a neural network that can transform $\texttt{Unif}(0,1)$ to an arbitrary distribution $P$ (cf. Lemma $5$ in \cite{10.5555/3540261.3541089}). To facilitate easier learning of NCMs, we assume $\mathcal{N}(0,1) \mapsto P(\widehat{U}), \forall~\widehat{U} \in \widehat{\mathbf{U}}$ in this work. Note that Lemma $5$ can be trivially extended to this case.


\textbf{Kernel Least Squares} It is well known that $\mathbb{E}[Y|X]=\textup{argmin}_{f}\ \mathbb{E}\left[\left\|f(X)-Y\right\|^2\right]$. When the kernel embeddings of $Y$ are used instead, it is known as kernels least squares, written as: $\mathbb{E}[\Phi(Y)|X]= \textup{argmin}_{f}\ \mathbb{E}\left[\left\|f(X)-\Phi(Y)\right\|^2\right]$, where $\Phi$ is the canonical feature map corresponding to a kernel. In case the kernel is characteristic (\cite{JMLR:v12:sriperumbudur11a}), $Y\mapsto\mathbb{E}[\Phi(Y)|X]$ is injective and characterizes the distribution of $Y$. Common examples of characteristic kernels include the radial basis function (\emph{RBF}) kernel and inverse multi-quadric kernel (\emph{IMQ}) kernel among others. Thus, kernel least- squares loss is well-suited for learning conditional distributions (e.g., see \cite{pmlr-v238-manupriya24a} which provides empirical and theoretical results in this regard). We espouse the following derivation from \cite{pmlr-v238-manupriya24a}, which shows how to learn a conditional generator using joint samples, without having to resort to Monte Carlo methods. Let $\mathcal{D}$ be a given dataset of samples drawn from the joint distribution of random variables $\mathrm{P}, \mathrm{Q}$ and let  $\pi_{\mathrm{Q}|\mathrm{P}}^{\gamma}$ be a parametrized (by $\gamma$) conditional generator that we wish to learn. Accordingly, we wish $\pi_{\mathrm{Q}|\mathrm{P}}^{\gamma}(\cdot|p) = \mathrm{s}_{\mathrm{Q}|\mathrm{P}}(\cdot|p), \forall p \in \Omega(\mathrm{P})$. Utilizing the injectivity of a characteristic kernel $\Phi$, we can equivalently rewrite the desired condition as $\int_{\Omega(\mathrm{P})}\Big\|\mathbb{E}_{\pi_{\mathrm{Q}|\mathrm{P}}^{\gamma}(\cdot|p)}[\Phi(Y)] - \mathbb{E}_{\mathrm{s}_{\mathrm{Q}|\mathrm{P}}(\cdot|p)}[\Phi(Y)]\Big\|_2^2\texttt{d}\mathrm{s}_{\mathrm{P}}(p) = 0$. The kernels least squares loss inside the above integral is commonly known as the squared Maximum Mean Discrepancy error, aka $\texttt{MMD}^2$. For the rest of this paper, we take $\Phi$ to be the \emph{IMQ} kernel defined as $k(x, y) = \frac{1}{\sqrt{\varrho + ||x-y||_2^2}} \forall x, y \in \mathbb{R}^d$, where $\varrho$ is a non-negative hyperparameter. This is because we usually observe good results with this kernel. 
Next, they apply a standard result kernel mean embeddings,
\cite{Muandet_2017}, which states that $\mathbb{E}[\|G - h(\mathrm{P})\|^2] = \mathbb{E}[\|G - \mathbb{E}[G|\mathrm{P}]\|^2] + \mathbb{E}[\|\mathbb{E}[G|\mathrm{P}]-\mathbb{E}[h(\mathrm{P})]\|^2]$, when $G$ is the kernel mean embedding of $\delta_{\mathrm{Q}}$ and $h(\mathrm{P})$ the kernel mean embedding of $\pi_{\mathrm{Q}|\mathrm{P}}^{\gamma}(\cdot|\mathrm{P})$. This helps us simplify the integral over the marginal of $\mathrm{P}$ in terms of a marginal over the joint distribution of $\mathrm{P}, \mathrm{Q}$, since $\int_{\Omega(\mathrm{P})}\texttt{MMD}^2(\pi_{\mathrm{Q}|\mathrm{P}}^{\gamma}(\cdot|p), \mathrm{s}_{\mathrm{Q}|\mathrm{P}}(\cdot|p))\texttt{d}\mathrm{s}_{\mathrm{P}}(p) + \vartheta(\mathrm{s}) = \int_{\Omega(\mathrm{P})\times\Omega(\mathrm{Q})}\texttt{MMD}^2(\pi_{\mathrm{Q}|\mathrm{P}}^{\gamma}(\cdot|p), \delta_{\mathrm{Q}})\texttt{d}\mathrm{s}_{\mathrm{P},\mathrm{Q}}(p,q)$, where $\vartheta(\mathrm{s}) \geq 0$ is purely a function of the dataset and therefore, does not affect the minima of the right hand side of the equation above. The left hand side integral can be readily estimated empirically, as $\frac{1}{|\mathcal{D}|}\sum_{i=1}^{|\mathcal{D}|}\big\|\frac{1}{\kappa}\sum_{\widetilde{q_i} \sim \pi_{\mathrm{Q}|\mathrm{P}}^{\gamma}(\cdot|p_i)}^{\kappa}\Phi(\widetilde{q_i}) - \Phi(q_i)\big\|^2$. Training the conditional generator over the dataset $\mathcal{D}$ thereby is tantamount to minimizing the previously mentioned empirical estimate with respect to the parameters, $\gamma$. This "trick" is repeatedly used in most of our losses.

\subsection{Prior Work}\label{subsec:prior_work}
In this section, we focus on works particularly geared towards counterfactual fairness. As mentioned earlier, \cite{10.5555/3294996.3295162} introduced counterfactual fairness in its rigorous form, and proposed a Markov Chain Monte Carlo (MCMC) based approach to learn appropriate counterfactual samples, when the mean cannot be computed analytically. Since they consider a synthetic dataset for which the counterfactual distribution can be computed exactly, they use a probabilistic programming language to assess the counterfactual fairness of models trained using their method. However, using a probabilistic programming language may not be always be feasible because the distributions of interest may not always be known exactly. Second, MCMC based approaches are typically prone to significant approximation errors and are usually applicable only for discrete variables. This is in contrast to our work which works for continuous, (in)sensitive variables as well as proposes a simpler measure for counterfactual fairness. Subsequently, a number of papers have studied the same problem and proposed better algorithms, at least under certain settings \cite{10.24963/ijcai.2024/504,10.1145/3580305.3599408,10.1007/s10994-022-06206-8,zuo2022counterfactual,10.5555/3367032.3367236}.
\cite{10.24963/ijcai.2024/504} proposes an Evidence Lower Bound (ELBO)-based method to learn for counterfactual fairness. However, it is still vulnerable to $\mathcal{L}_3$ inconsistency issues. \cite{10.1145/3580305.3599408} apparently aims to enforce counterfactual fairness by learning a loss $\mathcal{L}_c$ which measures the sample-wise distance between factual and counterfactual samples. Furthermore, though they aim to learn fair representations by minimizing the $\texttt{MMD}$ loss between fair representations, it is not shown if this explicitly captures counterfactual fairness between the predictands themselves while training. Additionally it is also vulnerable to $\mathcal{L}_3$ inconsistency. \cite{10.1007/s10994-022-06206-8} is a baseline approach that we adopt, closest to our setting cum method. As other works however, they also adopt a weak fairness metric (MSE) to measure for counterfactual fairness, which is not fully faithful to Definition \ref{def:ctfairness}. Their generation approach however considers an $\texttt{MMD}$ distance between priors and posteriors when computing the ELBO, coupled with an adversarial losses to generate sensitive values conditioned on inferred latents. Though it is claimed that this method achieves superior performance compared to learning with $\texttt{MMD}$ costs alone, their $\texttt{MMD}$-based variant critically defers from ours, as we sidestep an ELBO-based approach. 

Apart from the limitations highlighted above, to the best of our knowledge, no prior work has explicitly used the framework of NCMs to improve impartiality of a predictor to a certain sensitive attribute, which enhances the novelty of our methodology. 

\section{Counterfactually Fair Prediction}\label{sec:cfp}
As discussed in previous sections, we train a fair predictor in two stages. The first stage involves generating counterfactual data to aid the learning of the fair predictor, and the second stage involves training the fair predictor using a fairness metric on the dataset, augmented with counterfactual samples generated from the first stage. We begin by introducing our novel contributions in each of these stages incrementally.

\subsection{Improved Counterfactual Generation}\label{subsec:icg}
As in many prior works \cite{10.1007/s10994-022-06206-8,10.5555/3294996.3295162}, we require a causal graph that describes the generation of causal data, consisting of the sensitive attributes, protected attributes, the predictand and any other unobserved confounders of interest. In order to simplify the explanation of our method and compare our results to that of previous works, we stick to the causal data-generating graph depicted in Figure \ref{fig:causalgraph}. Nonetheless, our method can be instantiated for arbitrary causal graphs as well. 
\subsubsection{Modeling}\label{subsubsec:modelling}
We propose modeling causal mechanisms of interest using NCMs, because of their versatility and competence in executing counterfactually generative tasks \cite{xia2023neural}. In the context of Figure \ref{fig:causalgraph}, we model the causal mechanism $\mathcal{M}^{*}(\cdot, \cdot)$ that generates $X$ from $A$ and $U$ using a neural network $\widehat{\mathcal{M}_{\theta}}(\cdot, \cdot)$. Note that unlike random variables $A$ and $X$, samples from $U$ are not observed. To handle this, we again borrow a trick from NCMs \cite{10.5555/3540261.3541089}, where an equivalent causal model $\widehat{\cal{M}^{*}}(\cdot, \cdot)$ that generates $X$ from $A, \widehat{U}$, where $\widehat{U} \sim \mathcal{N}(0,\texttt{I}_{d\times d})$ is considered, and $\widehat{\mathcal{M}_{\theta}}(\cdot, \cdot)$ technically models $\widehat{\mathcal{M}^{*}}(\cdot, \cdot)$. However, our losses differ from \cite{10.5555/3540261.3541089} in that we train our models using kernels least squares losses between appropriate distributions whereas they train their models using negative log-likelihood.

Moreover, in order to generate counterfactual samples, we need to be able to abduct the exogenous variables, in accordance with Pearl's counterfactual inference recipe \cite{https://doi.org/10.1111/cogs.12065}. Specifically in the context of Definition \ref{def:ctfairness}, we require samples from $U_{|A=a, X=x}$, where $(A=a, X=x)$ is the evidence observed when generating a counterfactual sample of $X$, and its subsequent prediction along with the intervened sensitive attribute, $A\leftarrow a'$. Accordingly, we model the true abductor of the equivalent causal model $\widehat{\mathcal{M}^{*}}(\cdot, \cdot)$, $\widehat{\mathcal{A}^{*}}(\cdot, \cdot)$ that generates samples from $U_{|A=a, X=x}$ when given a sample from the joint distribution $(A,X)$ as evidence $e$, using a neural network $\widehat{\mathcal{A}_{\psi}}(\cdot, \cdot, \cdot)$, which also takes $e$ as input along with some samples from $\mathcal{N}(0,1)$ as pushforward noise. Unlike a couple of works that study counterfactual estimation and identification and assume invertability of the causal mechanisms \cite{10.5555/3618408.3619478,10.24963/ijcai.2024/907}, we do NOT require the same assumption. 

\subsubsection{Training}\label{subsubsec:training}
We begin by training $\widehat{\mathcal{M}_{\theta}}(\cdot, \cdot)$ to be $\mathcal{L}_1$ consistent (ref. \cite{10.5555/3540261.3541089}) w.r.t $\widehat{\mathcal{M}^{*}}(\cdot, \cdot)$ by using the observational data. The dataset is provided as a list of $n$ triplets, $\{(a_i, x_i, y_i)\}_{i=1}^{n}$, and the goal is to learn $\widehat{\mathcal{M}_{\theta}}(A, \cdot)$ such that $P_{\widehat{\mathcal{M}_{\theta}}}(A, \widehat{U}) = P_{\widehat{\mathcal{M}^{*}}}(A, \widehat{U})$. To this end, we critically employ the fact that any characteristic kernel $\Phi$ induces an injective map over distributions, stated in Section \ref{subsec:prelims}, as follows:

\begin{equation}
\begin{aligned}[t]
\label{eqn:gen_loss_derivation}
    P_{\widehat{\mathcal{M}_{\theta}}}(A, \widehat{U}) = P_{\widehat{\mathcal{M}^{*}}}(A, \widehat{U}) \Leftrightarrow ~\forall a \in \Omega(A), \\ P_{\widehat{\mathcal{M}_{\theta}}}(A=a, \widehat{U}) = P_{\widehat{\mathcal{M}^{*}}}(A=a, \widehat{U}) \Leftrightarrow ~\forall a \in \Omega(A), \\ \mathbb{E}_{X \sim P_{\widehat{\mathcal{M}_{\theta}}}(A=a, \widehat{U})}[\Phi(X)] = \mathbb{E}_{X \sim P_{\widehat{\mathcal{M}^{*}}}(A=a, \widehat{U})}[\Phi(X)]
\end{aligned}
\end{equation}

Consequently, we aim to learn a conditional generator $\widehat{\mathcal{M}_{\theta}}$, using joint samples $\{(a_i, x_i)\}_{i=1}^{n}$. This is our preferred loss to train over observational data due to its demonstrated superior performance compared to other traditional losses such as adversarial/KL/Wasserstein losses \cite{pmlr-v238-manupriya24a}.

In particular, we use the $\texttt{MMD}^2$ loss between the empirical means of the conditional distributions $P_{\widehat{\mathcal{M}_{\theta}}}( \widehat{U}|A=a)$ and $P_{\widehat{\mathcal{M}^{*}}}( \widehat{U}|A=a)$ as follows:

\begin{equation}\label{eqn:gen_loss}
\begin{aligned}[t]
    \ell_{\texttt{gen}} = \frac{1}{n_{\texttt{gen}}}\sum_{i=1}^{n_{\texttt{gen}}}\left\|\frac{1}{q_{\texttt{gen}}}\sum_{j=1}^{q_{\texttt{gen}}}\Phi(\widehat{\mathcal{M}_{\theta}}(a_i, \eta_{ij})) - \Phi(x_i)\right\|_2^2.
\end{aligned}
\end{equation}

Here, $n_{\texttt{gen}}$ is the number of samples within the training set (or within a mini-batch, if using an optimization algorithm such as mini-batch SGD \cite{}) and $q_{\texttt{gen}}$ the number of noise samples $\eta_{ij}$ sampled from $\widehat{U} \sim \mathcal{N}(0,I_{d \times d})$ per data point $(x_i, a_i)$, used to empirically approximate the kernel mean of $X$ when in turn sampled from $P_{\widehat{\mathcal{M}_{\theta}}}(\widehat{U}|A=a)$. For the sake of completeness, we also note that alternatives from conditional generative adversarial networks based literature may also be viable to learn this conditional generator, as we do not claim the above loss as a novel contribution of our work.

Enroute crafting $\widehat{\mathcal{M}_{\theta}}$ to be $\mathcal{L}_2$ \& $\mathcal{L}_3$ consistent w.r.t $\widehat{\mathcal{M}^{*}}$, we notice that since $X$ is Markovian with respect to $A$ in Figure \ref{fig:causalgraph}, the interventional distribution of $A$ degenerates to the conditional distribution of $A$ that we have already tackled by learning $\widehat{\mathcal{M}_{\theta}}$ via the loss in Equation \ref{eqn:gen_loss}. This is in line with many works in causal learning literature that assume Markovianity \cite{JMLR:v25:21-1440,10.5555/3618408.3619478}. We thus focus on making $\widehat{\mathcal{M}_{\theta}}$ $\mathcal{L}_3$ consistent w.r.t $\widehat{\mathcal{M}^{*}}$, which entails that we learn $\widehat{\mathcal{A}_{\psi}}$ that is distributionally identical to $\widehat{\mathcal{A}^{*}}$. In particular, we desire $P_{\widehat{\mathcal{M}^{*}}, \widehat{\mathcal{A}^{*}}}(X, A, U) = P_{\widehat{\mathcal{M}_{\theta}}, \widehat{\mathcal{A}_{\psi}}}(X, A, U)$. Unlike, Equation \ref{eqn:gen_loss} where we resorted to matching the conditional distributions, here we adopt matching the joint distributions directly, akin to the amortized implicit model described in \cite{NEURIPS2020_0987b8b3}. While matching conditional distributions is permissible here, we adopt a loss matching the joints primarily due to better empirical performance compared to the latter. Accordingly, we make the following observations:

\begin{equation}\label{eqn:pos_true_factor}
    \begin{aligned}[t]
        P_{\widehat{\mathcal{M}^{*}}, \widehat{\mathcal{A}^{*}}}(X, A, U) = P_{\widehat{\mathcal{M}^{*}}, \widehat{\mathcal{A}^{*}}}(X|A, U)P_{\widehat{\mathcal{M}^{*}}, \widehat{\mathcal{A}^{*}}}(A,U), \\ \Rightarrow P_{\widehat{\mathcal{M}^{*}},\widehat{\mathcal{A}^{*}}}(A,U) = P_{\widehat{\mathcal{M}^{*}}, \widehat{\mathcal{A}^{*}}}(A)P_{\widehat{\mathcal{M}^{*}}, \widehat{\mathcal{A}^{*}}}(U), \because A \perp \!\!\! \perp U, \\ \Rightarrow P_{\widehat{\mathcal{M}^{*}}, \widehat{\mathcal{A}^{*}}}(X, A, U) = P_{\widehat{\mathcal{M}^{*}}}(X|A, U)P(A)P(\widehat{U}).
    \end{aligned}
\end{equation}

The last line in Equation \ref{eqn:pos_true_factor} follows from the facts that $P_{\widehat{\mathcal{M}^{*}}, \widehat{\mathcal{A}^{*}}}(U) = P(\widehat{U})$ by the definition of $\widehat{\mathcal{M}^{*}}$ as it assumes the prior of the unobserved confounder $U$ to be $\mathcal{N}(0,I_{d \times d})$, while $P_{\widehat{\mathcal{M}^{*}}, \widehat{\mathcal{A}^{*}}}(A)$ is simply the prior of the sensitive attribute $A$, written simply as $A$. Since this prior distribution is unknown in general, we fallback on approximating it via the sample mean of $A$ in the dataset $\{(x_i, a_i)\}_{i=1}^{n}$. Further crucially note here, that contingent on $M$ achieving the minimal loss w.r.t Equation \ref{eqn:gen_loss}, the distribution induced by $\widehat{\mathcal{M}_{\theta^{*}}}$ ($\theta^{*}$ being an optimal set of parameters) becomes identical to that of $\widehat{\mathcal{M}^{*}}$ and can thus be used in place of $\widehat{\mathcal{M}^{*}}$.
 
Similarly, we appropriately factor $P_{\widehat{\mathcal{M}_{\theta}}, \widehat{\mathcal{A}_{\psi}}}(X, A, U)$ as:

\begin{equation}\label{eqn:pos_pred_factor}
    \begin{aligned}[t]
        P_{\widehat{\mathcal{M}_{\theta}}, \widehat{\mathcal{A}_{\psi}}}(X, A, U) = P_{\widehat{\mathcal{M}_{\theta}}, \widehat{\mathcal{A}_{\psi}}}(X, A)P_{ \widehat{\mathcal{A}_{\psi}}}(U|X, A).
    \end{aligned}
\end{equation}
Since we seek to explicitly learn $\widehat{\mathcal{A}_{\psi}}$ via the factorization in Equation \ref{eqn:pos_pred_factor}, we approximate the sample mean of $P_{\widehat{\mathcal{M}_{\theta}}, \widehat{\mathcal{A}_{\psi}}}(X, A)$ via the sample mean of the dataset, akin to the case for Equation \ref{eqn:pos_true_factor}. Finally, like in Equation \ref{eqn:gen_loss}, we employ the $\texttt{MMD}^2$ loss between the joint distributions as:

\begin{equation}\label{eqn:pos_loss}
\begin{aligned}[t]
    \ell_{pos} = \Bigg\|\frac{1}{n_{\texttt{pos}}}\sum_{i=1}^{n_{\texttt{pos}}}\frac{1}{q_{\texttt{pos}}}\sum_{j=1}^{q_{\texttt{pos}}}\Phi\left(\widehat{\mathcal{M}_{\theta}}\left(a_i, \eta_{ij}\right), a_i, \eta_{ij}\right)\\-\frac{1}{n_{\texttt{pos}}}\sum_{i=1}^{n_{\texttt{pos}}}\frac{1}{q_{\texttt{pos}}}\sum_{j=1}^{q_{\texttt{pos}}}\Phi\left(x_i, a_i, \widehat{\mathcal{A}_{\psi}}\left(x_i, a_i, \overline{\eta_{ij}}\right)\right)\Bigg\|_2^2.
\end{aligned}
\end{equation}
Other variants of Equation \ref{eqn:pos_loss} are also possible, such as using samples from $\widehat{\mathcal{M}_{\theta}}$ itself instead of samples $x_i$ from the dataset itself, but the underlying idea of matching samples from the joint distributions remains invariant. 

After training $\widehat{\mathcal{M}_{\theta}}, \widehat{\mathcal{A}_{\psi}}$ using losses Equations \ref{eqn:gen_loss} and \ref{eqn:pos_loss}, we are sufficiently equipped to generate counterfactual samples given an intervening value $a' \in \Omega(A)$, and evidence $(x, a) \in \Omega(X) \times \Omega(A)$, as $\widehat{\mathcal{M}_{\theta}}(a', \widehat{\mathcal{A}_{\psi}}(x, a, \overline{\eta}))$, where $\bar{\eta} \sim \mathcal{N}(0,I_{d \times d})$. As such, as proved in \cite{xia2023neural}, NCMs are expressive enough and are $\mathcal{L}_3$ consistent, as a family of causal models. Here, we make a vital observation that most of these results on $\mathcal{L}_3$ consistency hold assuming the posterior distribution is perfectly learned. However, for estimating the counterfactual, the posterior of the exogenous variables need to be estimated. In practice, this estimation may make the resultant model $\mathcal{L}_3$ inconsistent. This issue is ignored in this paper. To the best of our knowledge, this issue appears to have been overlooked in many works \cite{xia2023neural,10.5555/3540261.3541089,Xia_Bareinboim_2024,NEURIPS2023_65cbe3e2,liu2024identifiable,9578520}, specifically in the context of NCMs. Secondly, a closer look at arguably one of the pioneering works in this direction \cite{xia2023neural}, suggests that the Monte-Carlo based estimation proposed therein is specific to discrete causal variables. Thirdly, in the case of non-identifiablity the Algorithm $3$ in \cite{xia2023neural} does not provide an alternative.

Therefore, we seek to alleviate these issues and thus propose a more widely-applicable counterfactual generator by introducing a novel loss, that we term $\ell_{\texttt{ctf}}$. Since the neural regressor(s) trained via Equations \ref{eqn:gen_loss} \& \ref{eqn:pos_loss} may spoil the $\mathcal{L}_3$ consistency, a key idea is to impose the $\mathcal{L}_3$ consistency conditions explicitly, thereby encouraging the counterfactual distributions implicitly learned to mirror the marginal law over counterfactuals, described in Section \ref{subsec:prelims}. For example, in the exogenous case ($A \rightarrow B$), this boils down to Equation $19$ in \cite{probCause}, or equivalently Proposition $10$ in \cite{JMLR:v25:21-1440}.
In particular, we seek to enforce:

\begin{equation}\label{eqn:ctf_loss_derive}
    \begin{aligned}[t]
        \sum_{(a, x)}P_{\widehat{\mathcal{M}_{\theta}}}(A\leftarrow a', \widehat{\mathcal{A}_{\psi}}(A=a, X=x))P_{}(A=a, X=x) = \\P_{\widehat{\mathcal{M}^{*}}}(A\leftarrow a', \cdot) = P_{\widehat{\mathcal{M}^{*}}}(A=a', \cdot), (a,x) \in\Omega(A)\times\Omega(X).
    \end{aligned}
\end{equation}

Appropriately employing the $\texttt{MMD}^2$ loss then gives:

\begin{equation}\label{eqn:ctf_loss}
    \begin{aligned}[t]
        \ell_{\texttt{ctf}} = \frac{1}{n_{\texttt{ctf}}}\sum_{i=1}^{n_{\texttt{ctf}}}\left\|\frac{\mathcal{C}}{n_{\texttt{ctf}}q_{\texttt{ctf}}} - \Phi(x_i)\right\|^2, \text{where}~\\
        \mathcal{C} = \sum_{j=1}^{n_{\texttt{ctf}}}\sum_{k=1}^{q_{\texttt{ctf}}}\Phi\left(\widehat{\mathcal{M}_{\theta}}\left(a_i, \widehat{\mathcal{A}_{\psi}}\left(\widehat{\mathcal{M}_{\theta}}\left(a_j, \eta_{ijk}\right), a_j, \overline{\eta_{ijk}}\right)\right)\right).
    \end{aligned}
\end{equation}

Notice how training $\widehat{\mathcal{M}_{\theta}}$ using Equation \ref{eqn:ctf_loss} aims to mitigate the issues of $\mathcal{L}_3$ inconsistency due to inconsistent posterior estimation, as well as cleverly avoids using Monte Carlo estimates by making use of kernels least squares loss. However, this alone does not handle scenarios where the counterfactual distribution is inherently non-identifiable. In particular, there could be many NCMs $\widehat{\mathcal{M}_{\theta^{*}}}$ that achieve optimally low loss on all three losses combined, i.e. Equations \ref{eqn:gen_loss}, \ref{eqn:pos_loss} \& \ref{eqn:ctf_loss}. 
Thus, we can optionally induce a bias in our counterfactual generator model that outputs \emph{near-world} counterfactuals \cite{Wachter2017CounterfactualEW}. We can accordingly employ an OT-based regularizer as:

\begin{equation}\label{eqn:reg_loss}
    \begin{aligned}[t]
        \ell_{\texttt{reg}} = \frac{1}{n_{\texttt{reg}}^2}\sum_{i=1}^{n_{\texttt{reg}}}\sum_{j=1}^{n_{\texttt{reg}}}\texttt{dist}(\widehat{\mathcal{M}_{\theta}}(a_i, \widehat{\mathcal{A}_{\psi}}(x_j, a_j, \overline{\eta_j})), x_j),
    \end{aligned}
\end{equation}

where $\texttt{dist}$ is an appropriate distance metric, such as Euclidean, Wasserstein, etc. In this work, we assume the distance metric as Euclidean for simplicity. Also note that unlike the other losses, $\ell_{\texttt{reg}}$ is usually chosen to be a pointwise loss, in that it doesn't necessarily explicitly minimize the distance between distributions, but rather the distance between the actual samples from the distributions. 

Finally, we can jointly train $\widehat{\mathcal{M}_{\theta}},\widehat{\mathcal{A}_{\psi}}$ using all the losses combined, in which case the optimization problem becomes:
\begin{equation}\label{eqn:joint_loss}
\begin{aligned}[t]
\min_{\theta,\psi}\lambda_{\texttt{gen}}\ell_{\texttt{gen}}(\theta)+\lambda_{\texttt{pos}}\ell_{\texttt{pos}}(\theta,\psi)\\+\lambda_{\texttt{ctf}}\ell_{\texttt{ctf}}(\theta,\psi)+\lambda_{\texttt{reg}}\ell_{\texttt{reg}}(\theta,\psi),
\end{aligned}
\end{equation}
as well as train them separately, in which case the optimization problem becomes:
\begin{equation}\label{eqn:phased_loss}
\begin{aligned}[t]
\theta^{*} \equiv \arg\min_{\theta}\ell_{\texttt{gen}}(\theta) \\
\psi^{*} \equiv \arg\min_{\psi} \lambda_{\texttt{pos}}\ell_{\texttt{pos}}(\theta^{*},\psi)+\\\lambda_{\texttt{ctf}}\ell_{\texttt{ctf}}(\theta^{*},\psi)+\lambda_{\texttt{reg}}\ell_{\texttt{reg}}(\theta^{*},\psi).
\end{aligned}
\end{equation}

\subsection{Fairness Finetuning}\label{subsec:ff}
In the second stage, we aim to learn a counterfactually fair predictor. Reminiscent of \cite{10.1007/s10994-022-06206-8}, we train our predictor $h_{\phi}(\cdot, \cdot)$ with a joint loss of the form:

\begin{equation}\label{eqn:joint_fair_loss}
    \begin{aligned}[t]
        \ell_{\texttt{pred}} + \lambda_{\texttt{fair}}\ell_{\texttt{fair}},
    \end{aligned}
\end{equation}
where $\ell_{\texttt{pred}}$ is a standard supervised loss over the dataset, e.g. mean squared error (MSE) in the case of regression, or cross entropy (CE) in the case of classification, and $\ell_{\texttt{fair}}$ is a fairness metric based off Definition \ref{def:ctfairness}. Our novel insight lies in employing an $\texttt{MMD}^2$ loss between the kernel embedding means of the distributions as the fairness metric, since the metric returns $0$ precisely when the distributions on the right and left hand sides of Definition \ref{def:ctfairness} are identical, and returns a non-zero loss value otherwise. To the best of our knowledge, this is surprisingly contrasted by a number of prior works in the domain of counterfactual fairness \cite{10.1007/s10994-022-06206-8,10.1145/3580305.3599408,10.5555/3294996.3295162,zuo2022counterfactual} which test for distributional equality via a weak metric such as MSE/RMSE between the sample means of the distributions, which may be $0$ even when the distributions are not identical (as a simple example, consider the two different distributions $\mathcal{N}(0, 1)$ \& $\mathcal{N}(0, 2)$ that have identical means). 
Specifically, our fairness metric $\ell_{\texttt{fair}}$ reads as:

\begin{equation}\label{eqn:fairness_metric}
    \begin{aligned}[t]
         \frac{1}{n_{\texttt{fair}}q_{\texttt{intv}}}\sum_{i=1}^{n_{\texttt{fair}}}\sum_{j=1}^{q_{\texttt{intv}}}\Bigg\|\frac{1}{q_{\texttt{abd}}}\sum_{k=1}^{q_{\texttt{abd}}}\mathcal{F}^{\texttt{ctf}}_{ijk}-\frac{1}{q_{\texttt{abd}}}\sum_{k=1}^{q_{\texttt{abd}}}\mathcal{F}^{\texttt{fact}}_{ijk}\Bigg\|_2^2,\\ \text{where}~
        \mathcal{F}^{\texttt{ctf}}_{ijk} = \Phi\left(h_{\phi}\left(\widehat{\mathcal{M}_{\theta}}\left(a_j,\widehat{\mathcal{A}_{\psi}}\left(x_i, a_i, \widetilde{\eta_{ijk}}\right)\right), a_j\right)\right), \\ \text{and}~
        \mathcal{F}^{\texttt{fact}}_{ijk} = \Phi\left(h_{\phi}\left(\widehat{\mathcal{M}_{\theta}}\left(a_i,\widehat{\mathcal{A}_{\psi}}\left(x_i, a_i, \overline{\eta_{ijk}}\right)\right), a_i\right)\right).
    \end{aligned}
\end{equation}
Analogous to Equations \ref{eqn:gen_loss} \& \ref{eqn:ctf_loss}, Equation \ref{eqn:fairness_metric} is also a conditional loss, which measures the expected counterfactual fairness conditioned on $n_{\texttt{fair}}$ samples from the dataset, $\{x_i, a_i\}_{i=1}^{n_{\texttt{fair}}}$. For each pair $(x_i, a_i)$, we randomly sample $q_{\texttt{intv}}$ interventional values for the sensitive attribute $A$ as $\{a_j\}_{j=1}^{q_{\texttt{intv}}}, a_j \in \Omega(A)$. Thus we form $n_{\texttt{fair}}q_{\texttt{intv}}$ triplets of the form $(x_i, a_i, a_j)$, and for each such triplet, we compute the kernel mean squared discrepancy between counterfactual distributions induced by $a_j, a_i$ respectively, where each mean is computed over $q_{\texttt{abd}}$ samples. Further note that do not approximate the counterfactual distribution induced by abducting the posterior exerted by $(x_i, a_i)$, and intervening back with $a_i$ by $x_i$, since the variance inherent to the posterior may generate samples that are not identical to $x_i$.

We use $\ell_{\texttt{fair}}$ as a loss to enforce counterfactual fairness during training, as well as propose using it as measure of counterfactual fairness of the predictor over the test set. Selected prior works such as \cite{10.1007/s10994-022-06206-8,10.1145/3580305.3599408} report \texttt{MMD} scores between distributions of certain representations typically using \emph{RBF} kernel (which is a characteristic kernel), but still fall shy of reporting the scores between the actual distributions of interest as per Definition \ref{def:ctfairness}. Unlike $\texttt{MMD}$ which is a metric, $\texttt{MMD}^2$ is NOT a metric over the space of probability distributions. However we prefer training with $\texttt{MMD}^2$ loss rather than $\texttt{MMD}$ loss since the former is smooth and strongly convex, while the latter isn't. Thus it is easier to optimize using $\texttt{MMD}^2$ loss, explaining our preference. Thus, we propose using Equation \ref{eqn:fairness_metric} as an appropriate fairness metric to measure the extent of counterfactual fairness. 

\section{Experiments}\label{sec:experiments}
We empirically evaluate our proposed methodology on two datasets, referring \cite{10.1007/s10994-022-06206-8}. Since we primarily focus on continuous attributes in this work, we choose the synthetic Insurance dataset \cite{10.1007/s10994-022-06206-8} and the real world Crimes dataset \cite{communities_and_crime_183}. 
\subsection{Datasets}\label{subsec:datasets}
The synthetic \textbf{Insurance} dataset consists of a five-dimensional confounder $\mathbf{U}$, coupled with a uni- dimensional sensitive attribute $A$; $\mathbf{X}, Y$ are four-dimensional and uni-dimensional respectively (for exact details on this dataset, we kindly refer the reader to \cite{10.1007/s10994-022-06206-8}). Setting $U$ explicitly, as a multivariate normal with diagonal covariance matrix helps to analytically compute the exact counterfactual distributions for reference. In particular, it involves reparametrizing $Y$ and then solving a system of linear equations involving $\mathbf{X}, A$ to compute the posterior distribution of $\mathbf{U}$. We normalize the data following \cite{10.1007/s10994-022-06206-8}, and train using a dataset size of $5000$ samples and incorporate a train-test split of $80/20$ respectively. After training for counterfactual generation once, we train for counterfactual fairness with multiple different values of lambda, where we repeat training and testing for each value of lambda between $3-5$ times, to marginalize over all sources of randomness when performing the experiments. 

The \textbf{Crimes} dataset on the other hand is even higher dimensional, since $\textbf{X}$ is effectively $121$-dimensional here (dimensions corresponding to many missing/inappropriate values are dropped from the dataset). $X, Y$ on the other hand are unidimensional. We adopt a $~90/10$ train-test split for this dataset, with $1794$ samples used for training and $200$ samples for testing. Akin to Insurance, we repeat fairness training for each lambda multiple times in reporting the results. 

For compatibility, we use the exact same architecture as \cite{10.1007/s10994-022-06206-8} for $h_{\phi}$, while for $\widehat{\mathcal{M}_{\theta}}$ and $\widehat{\mathcal{A}_{\psi}}$ we use 3-layered neural networks, with each layer having hidden dimension $32$. We experimented with multiple learning rates and hyperparameter values for the \emph{IMQ} kernel and report the best values in the figures\footnote{Code will be released upon acceptance}. We emphasize that we only compare our method on datasets where all attributes are continuous to better bring out the efficacy of our method as well as to simplify the presentation by sticking to the \emph{IMQ} kernel. Adapting our method to discrete data is possible with a few caveats, such as changing the kernel to a $0-1$ kernel. In particular for sampling intervening values from $\Omega(A)$, we simply sample from a computationally convenient distribution like $\mathcal{N}(0,1)$ as in both these datasets, the domain $\Omega(A) \subseteq \mathbb{R}$.
\subsection{Evaluation}\label{subsec:eval}
As mentioned in the introduction and backed by results in \cite{10.1007/s10994-022-06206-8}, the performance of a predictor (e.g. accuracy in the discrete case, MSE in the continuous case) is usually inversely proportional to the extent to which it is counterfactually fair. This phenomenon can be intuitively understood by examining the expected behavior of the prospective fair predictor at two extremes, aka when $\lambda = 0$ and $\lambda \rightarrow \infty$ in Equation \ref{eqn:joint_fair_loss}. When $\lambda \rightarrow \infty$, the supervised loss $\ell_{\texttt{pred}}$ is practically rendered otiose, since $h_{\phi}(\cdot, \cdot)$ can trivially learn a constant function and achieve the optimal counterfactual fairness loss, infact equal to $0$. However the performance of such a predictor on the dataset will expectantly be unacceptable. On the other hand, when $\lambda = 0$, an optimal predictor $h_{\phi*}(\cdot, \cdot)$ perfectly represents the dataset $\{(x_i, a_i, y_i)\}_{i=1}^{n}$. However, it absorbs all the inequity present in the dataset itself and is thus expected to have low counterfactual fairness score compared to when $\lambda \rightarrow \infty$. 

Since comparing for counterfactual fairness across methods for a specific data point $(a, x, y)$ may involve tuning for method-intrinsic hyperparameters and sample-approximation issues, we propose a more generic way of measuring and comparing counterfactual fairness across methods. Let $\Gamma_1$ and $\Gamma_2$ be two methods that aim to learn counterfactually fair predictor. Suppose that we plot the performance ($\mathrm{E}$) against the counterfactual fairness of the predictor ($\mathrm{F}$), on the x- and y- axes respectively. Further assume that $\mathrm{E}$ is an increasing function of its performance (e.g. explained variance score is higher if the performance is better).

Intuitively then, we would called $\Gamma_1$ better poised to learn counterfactually fair models compared to $\Gamma_2$ if for each value of $\mathrm{E}=e$, $\Gamma_1$ returned a lower fairness score (aka better fairness) than $\Gamma_2$, i.e. $\mathrm{F}_{\Gamma_1}(\mathrm{E}=e) \leq \mathrm{F}_{\Gamma_2}(\mathrm{E}=e)$. Since it may not feasible to expect either method to beat the other for every value of $\mathrm{E}=e$, we hope for the same to happen at least on average. We can mathematically write this as:

\begin{equation}\label{eqn:fair_auc}
    \begin{aligned}[t]
        \int \left(\mathrm{F}_{\Gamma_2}(\mathrm{E}) - \mathrm{F}_{\Gamma_2}(\mathrm{E})\right)\mathrm{d}\mathrm{E} \geq 0 \\ \implies \int \mathrm{F}_{\Gamma_2}(\mathrm{E})\mathrm{d}\mathrm{E} \geq \int\mathrm{F}_{\Gamma_1}(\mathrm{E})\mathrm{d}\mathrm{E}.
    \end{aligned}
\end{equation}

Notice that $\int\mathrm{F}_{\Gamma_i}(\mathrm{E})\mathrm{d}\mathrm{E}$ is the area under the curve (AUC) in the plot for method $\Gamma_i$. Thus verifying iff $\Gamma_1$ is superior to $\Gamma_2$ reduces to deducing whether the AUC for $\Gamma_1$ is atmost that of $\Gamma_2$. Switching the axes or increasing/decreasing nature of $\mathrm{E}$ switches the inequality. We adopt this method to compare our method against other approaches.

\subsection{Analysis}\label{subsec:analysis}
On close inspection, Figure \ref{fig:crimesplot} clearly indicates that the plot of \textbf{\texttt{our\_m\_our\_l}} has a higher AUC than \textbf{\texttt{our\_m\_their\_l}}, where the baseline is \cite{10.1007/s10994-022-06206-8}. The lines themselves are linear regressors of the different $(\mathrm{E}, \mathrm{F})$ points plotted for each method. Note that higher AUC indicates that our method is better, since the axes are switched here.
\begin{figure} \centering \includegraphics[width=\linewidth]{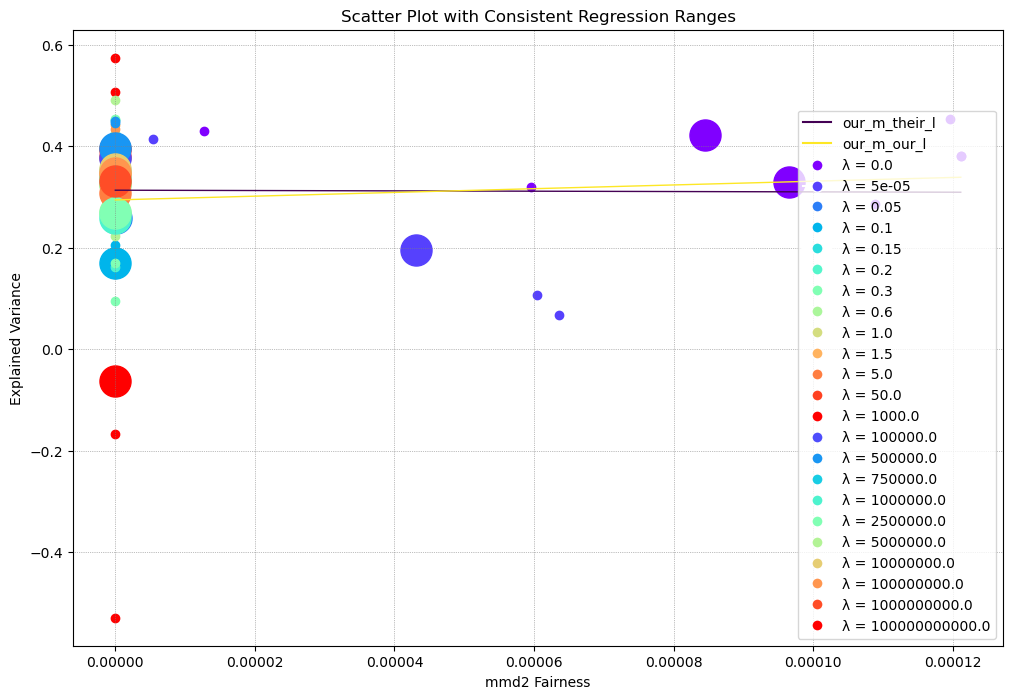} \caption{The legend \textbf{\texttt{our\_m\_{\{\}}\_l}} conveys that the respective line was observed by running our counterfactual generation method using either our fairness metric or the baseline's metric. The values of lambda indicate the different values at which the methods were tested.} \end{figure}\label{fig:crimesplot}

\begin{table*}
  \centering
  \begin{tabular}{|c*{9}{|c}|}
  \hline
     \multicolumn{1}{|c|}{\multirow{2}{*}{ \backslashbox{Method}{Dataset}}}
    & \multicolumn{3}{|c}{Synthetic} & \multicolumn{3}{|c|}{Crime}\\
    \cline{2-7}
     & \multicolumn{1}{|c|}{$\ell_{\texttt{ctf}}$} & \multicolumn{2}{|c|}{W\textbackslash o  $\ell_{\texttt{ctf}}$} & \multicolumn{1}{|c|}{$\ell_{\texttt{ctf}}$} & \multicolumn{2}{|c|}{W\textbackslash o $\ell_{\texttt{ctf}}$} \\
    \hline
    \cite{10.1007/s10994-022-06206-8} & 1.07 & 1.08  & 0.11 & 0 & 5 \\
    \hline 
  \end{tabular}
  \caption{Here is a caption}\label{tlc}
\end{table*}


\section{Conclusion}
The primary objective of this work is to move towards improved learning of counterfactually fair models. Existing literature in this regard largely suggests a common blueprint for enforcing such fairness by first learning counterfactual data/representations that can be later utilized in enforcing the empirical instantiation of the fairness metric employed. We identify that in doing so, most existing methods suffer from two problems, the first involving generation of partially fidel counterfactual samples due to potential $\mathcal{L}_3$ inconsistency induced due to errors in estimating the posterior distribution, and second of using weak metrics that do not capture the notion of counterfactual fairness exactly. We tackle the first issue by proposing two novel losses, grounded in the framework of NCMs, and the second by propounding an $\texttt{MMD}^2$-based metric. We show improved results for counterfactual fairness using a combination of these two ideas, demonstrating their advantage. For future directions, we note that enforcing $\mathcal{L}_3$ inconsistency may be of independent interest to the NCM community and suggest further exploring whether incorporating such a loss leads to improved counterfactual sample consistency, both empirically and theoretically \cite{zhou2024counterfactual}.

\bibliographystyle{named}
\bibliography{ijcai25}

\end{document}